\pdfoutput=1

\documentclass[11pt]{article}
\usepackage{naacl2021}

\usepackage[utf8]{inputenc} 
\usepackage[T1]{fontenc}    
\usepackage{hyperref}       
\usepackage{url}            
\usepackage{booktabs}       
\usepackage{amsfonts}       
\usepackage{nicefrac}       
\usepackage{microtype}      
\usepackage{url}
\usepackage{graphicx}
\usepackage{subcaption}
\usepackage{comment}

\usepackage{times}  
\usepackage{helvet} 
\usepackage{courier}  
\usepackage{xcolor}
\usepackage{amsmath} 
\usepackage{mathtools}
\usepackage{cancel}
\usepackage{multicol}
\usepackage{multirow}
\usepackage{booktabs}
\usepackage{wrapfig}
\usepackage{times}
\usepackage{latexsym}
\usepackage{amsfonts}  
\usepackage{nicefrac}  

\usepackage[T1]{fontenc}

\usepackage[utf8]{inputenc}

\usepackage{microtype}

\title{Privacy Regularization: Joint Privacy-Utility Optimization in Language Models}

\author{Fatemehsadat Mireshghallah\textsuperscript{\rm 1}\thanks{\quad Work done as part of an MSR internship.}, Huseyin A. Inan\textsuperscript{\rm 3}, Marcello Hasegawa\textsuperscript{\rm 2},\\
    \textbf{Victor Rühle}\textsuperscript{\rm 2}, \textbf{Taylor Berg-Kirkpatrick}\textsuperscript{\rm 1}, \textbf{Robert Sim}\textsuperscript{\rm 3}\\
    \textsuperscript{\rm 1} University of California San Diego,
    \textsuperscript{\rm 2} Microsoft Corporation, 
    \textsuperscript{\rm 3} Microsoft Research \\
    \texttt{\{fatemeh, tberg\}@ucsd.edu},\\ \texttt{ \{huseyin.inan, marcellh, virueh, rsim\}@microsoft.com}\\}

\begin{document}
\maketitle
\begin{abstract}
Neural language models are known to have a high capacity for memorization of training samples. This may have serious privacy implications when training models on user content such as email correspondence. 
Differential privacy (DP), a popular choice to train models with privacy guarantees, comes with significant costs in terms of utility degradation and disparate impact on subgroups of users.
In this work, 
%
%
we introduce two privacy-preserving regularization methods for training language models that enable joint optimization of utility and privacy through (1) the use of a discriminator and (2) the inclusion of a novel triplet-loss term.
We compare our methods with DP through extensive evaluation.
We show the advantages of our regularizers with favorable utility-privacy trade-off, faster training with the ability to tap into existing optimization approaches, and ensuring uniform treatment of under-represented subgroups.
\end{abstract}

\section{Introduction}
\vspace{-1ex}
\begin{figure*}
    \centering
     \includegraphics[width=0.9\linewidth]{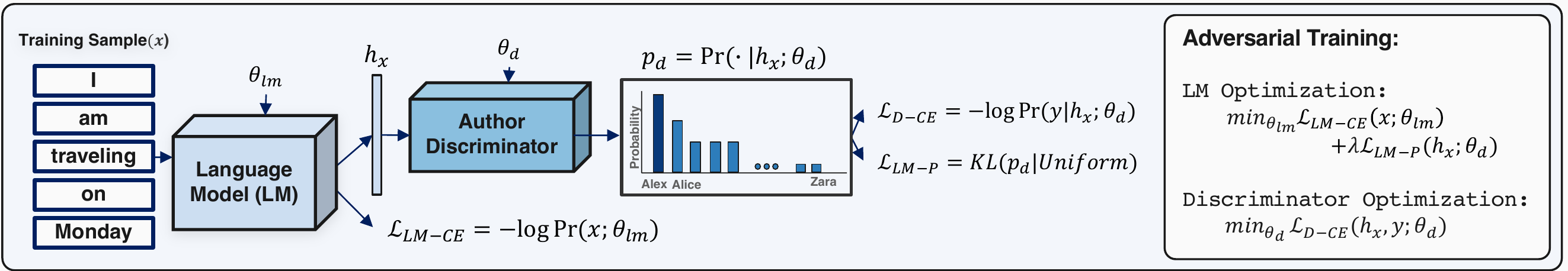}
     \caption{Workflow of our adversarial training regularization. The last hidden state ($h_x$) of the LM is fed to the discriminator to generate a distribution over the authors ($p_d$). $p_d$ is used to compute $\mathcal{L}_{\textsc{LM-P}}$, the privacy loss. 
     }
     \vspace{-1ex}
    \label{fig:prop}
    \vspace{-1ex}
\end{figure*}
Neural language models~\cite{Bengio2003ANP, mikolov10} have recently seen significant gains in capabilities, and are deployed at scale in several real-world scenarios
~\cite{chen2019gmail,adam2020ai}.
%
Training these models on domain-specific user data can further improve their utility.
The volume of data required, coupled with the inherent sparsity of natural language which often means all data are unique,
opens the door to an array of privacy attacks against models and their training data.  
%
Sample memorization poses a substantial risk by enabling  model inversion attacks~\cite{carlini2020extracting, ramaswamy2020training, inan2021training}. In these attacks, a curious or malevolent user can query a pre-trained language model on any data record with the intention of reconstructing (parts of) training samples\footnote{A  na\"ive example is an attacker querying ``My account number is'' and hoping to receive a user's account number.}. 
%
%
%

%
Differential Privacy (DP)~\cite{dwork2011differential} is the gold standard approach to address this issue, thanks to its strong and rigorous privacy guarantees.
DP-SGD ~\cite{abadi2016deep} is a popular method to train neural models with differential privacy guarantees and it works by clipping of the gradients and addition of noise in each update, which provides worst-case guarantees that reflect the likelihood of leaking any attribute of any member of the dataset into the trained model.
%
%
%
The worst-case guarantees of differential privacy are not customizable, in other words, they cannot be relaxed to protect only certain attributes. Therefore, DP incurs significant loss to model utility~\cite{tramer2020}. DP training of models is also much slower, with cumbersome hyper-parameter tuning and development~\cite{wu17, subramani2020enabling}.
%
%
%
%
It has also been shown that DP's utility loss is much worse for under-represented groups
~\cite{disparate, farrand2020}, which can have financial and societal ramifications~\cite{pujol2020}.
%
%

%
To address these issues,
we relax the strong assumptions of the DP threat model and assume an adversary with finite-capacity (finite statistical, compute, and side information) who attempts to recover sensitive user-level information from the trained model~\cite{carlini19}. 
%
We propose two privacy regularization methods, one based on adversarial training and another on a novel privacy loss term, to jointly optimize for privacy and utility of language models.
%
The main idea of our regularizers is to prevent the last hidden state representation of the language model
for an input sequence $x$ from being linked back to the sensitive attribute we are trying to protect, in our case, the identity of the author. We use the last hidden state as it corresponds to the embedding of the sequence $x$.\footnote{Although we consider recurrent neural network-based language models in this work, our approach is applicable in transformer-based language models as well. In the latter, one can consider the representation corresponding to the special token [CLS] as the embedding of the sequence $x$.}
We consider the linkability of the input representation to the sensitive attribute (author) as a proxy since it is commensurate with the linked and linkable information definitions in the General Data Protection Regulation~\cite{article29anon}. 
%
%
By framing privacy as an optimization problem, we can apply the well-developed machinery of large-scale gradient-based optimization, enabling us to train models at scale while jointly tuning for an optimal privacy-utility trade-off.

To validate our approach, we develop an evaluation framework for assessing a model's privacy loss. 
We employ the exposure metric introduced in \cite{carlini19} and introduce a reconstruction ({tab}) attack as a realistic scenario to evaluate and compare LSTM language models trained using our regularization with those trained with differential privacy, on Avocado~\cite{avocado2015} and Reddit~\cite{volske-etal-2017-tl} datasets. 
We also empirically demonstrate that, unlike DP, our technique does not have disparate impacts on under-represented groups. 
Our work is closely related to \cite{coavoux18} and \cite{li2018}.  \citeauthor{coavoux18} consider 
an attacker who eavesdrops on the hidden representations of a pre-trained model during inference and tries to recover information about the input text. Adversarial training is used as a mitigation to reduce the attacker's performance~\cite{wang2019balanced}. \citeauthor{li2018} use adversarial training to protect private author attributes such as age or gender, in learned text representations for part-of-speech tagging and sentiment analysis to gain better performance on out-of-domain corpora.
We, on the other hand, use adversarial training and a triplet-based regularization to train private language models that do not memorize sensitive user information, which has not been explored before. We evaluate our models accordingly, by trying to extract training samples. 
%
Prior work has studied membership inference attacks against models \cite{shokri17, yeom18, song19}, however, our regularizations do not target these attacks. 
\vspace{-1ex}
\section{Approach}
\vspace{-1ex}
In this section, we explain our proposed regularizers and training techniques in  more detail.

\begin{figure*}
    \centering
    \begin{subfigure}{0.3\textwidth}
     \centering
     \includegraphics[width=\linewidth, height=3cm]{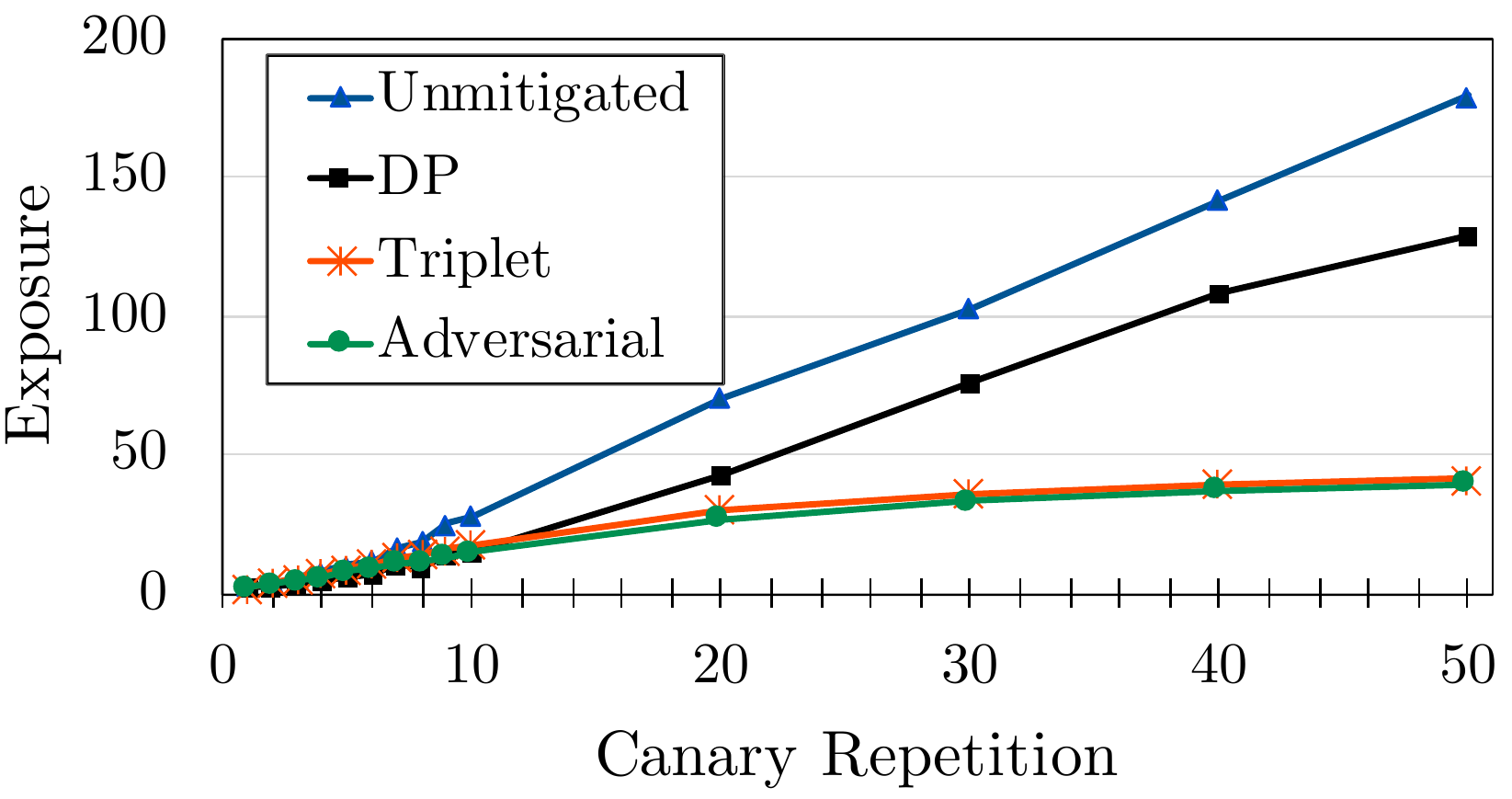}
     \vspace{-3ex}
     \caption{Avocado - High PPL ($\sim100$)}
     \label{fig:exposure:av-hi}
    \end{subfigure}
    \begin{subfigure}{0.3\textwidth}
    \centering
     \includegraphics[width=\linewidth, height=3cm]{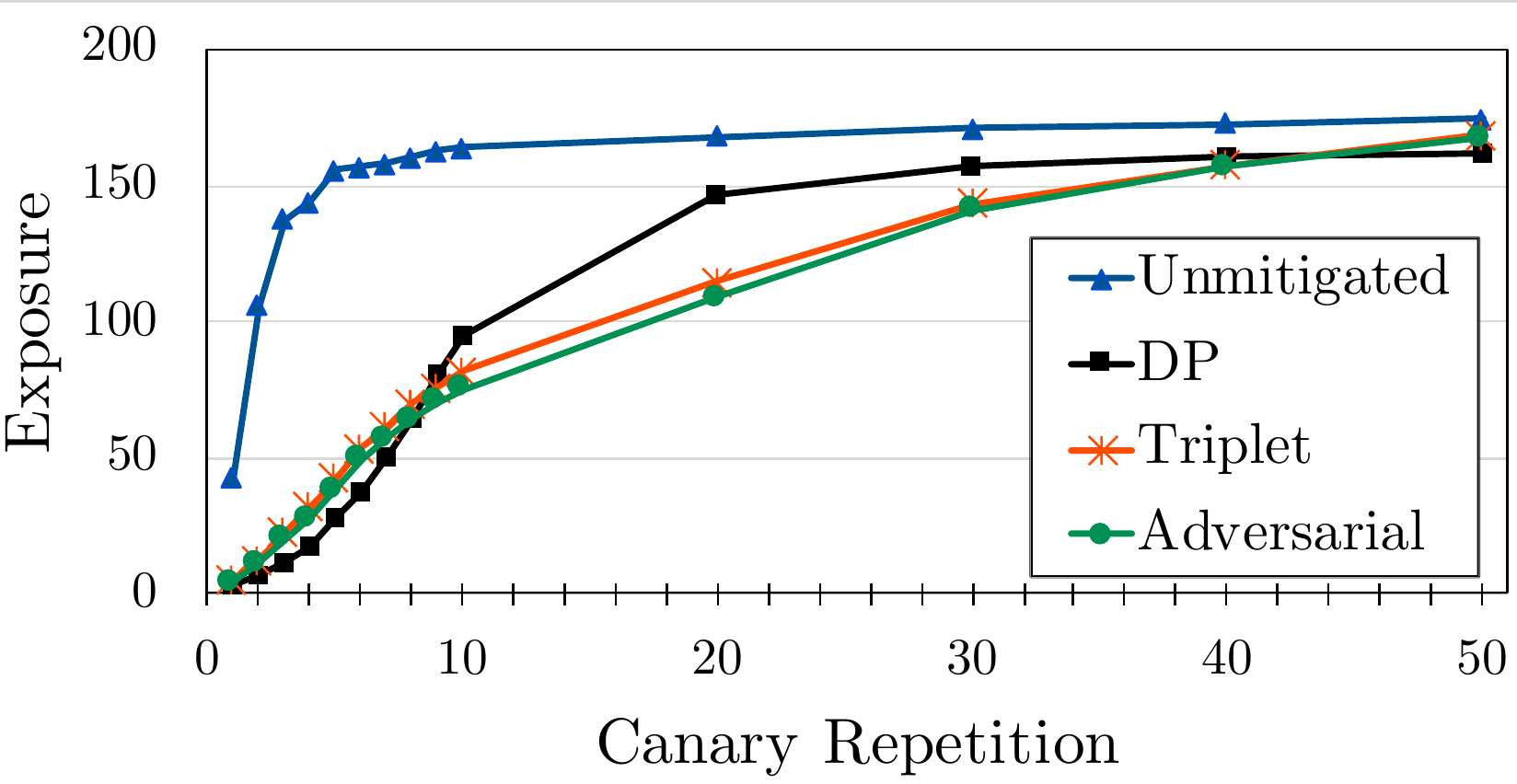}
     \vspace{-3ex}
     \caption{Avocado - Low PPL ($\sim60$)}
     \label{fig:exposure:av-lo}
    \end{subfigure}
    \begin{subfigure}{0.3\textwidth}
    \centering
     \includegraphics[width=\linewidth, height=3cm]{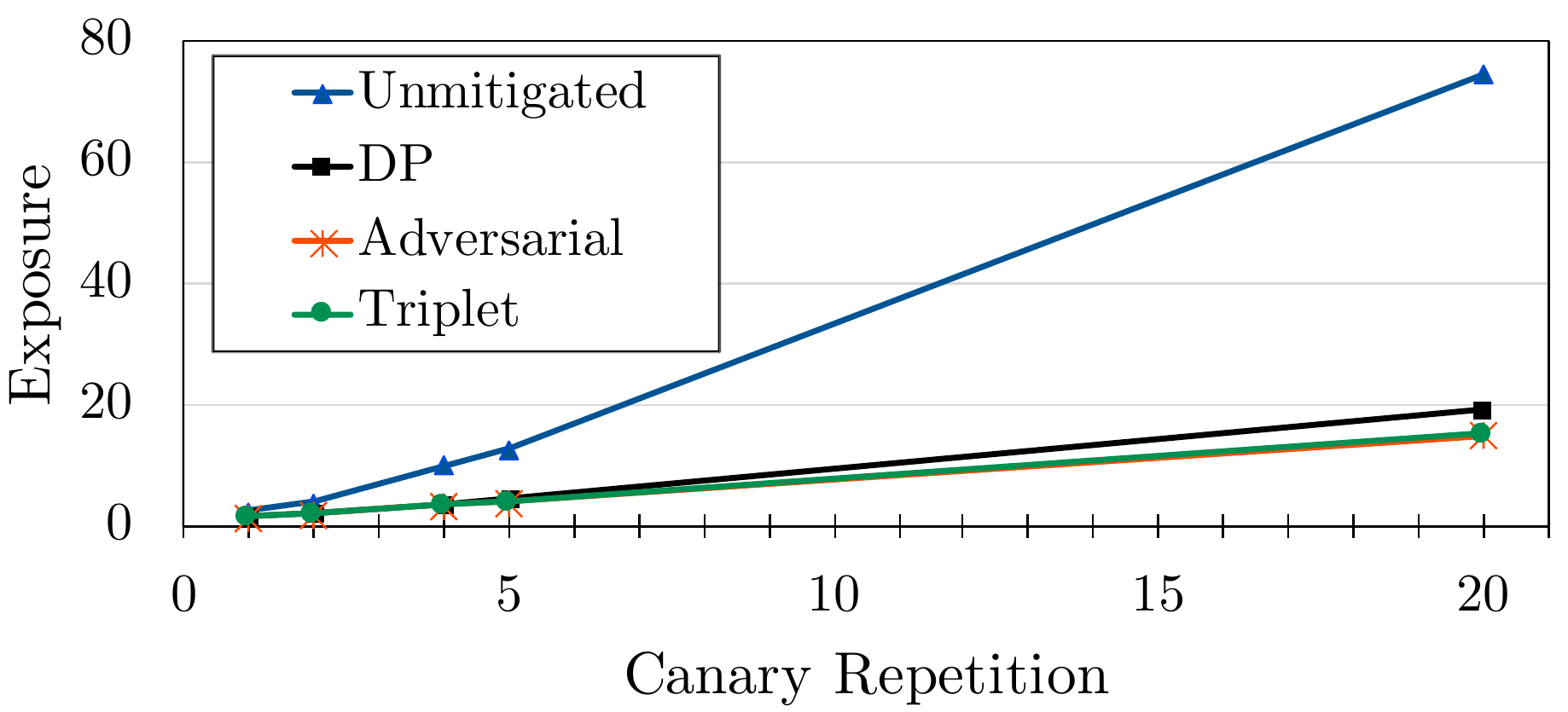}
     \vspace{-3ex}
     \caption{Reddit (PPL $\sim100$)}
     \label{fig:exposure:reddit}
    \end{subfigure}
    \caption{Exposure metric results for different training schemes at similar perplexities. Unmitigated denotes conventional training. Adversarial and Triplet  are our regularizers.
    Higher exposure indicates lower privacy.
    %
    %
    }
    \vspace{-2ex}
    \label{fig:exposure}
\end{figure*}


\subsection{Adversarial Training}

Figure~\ref{fig:prop} shows our first proposed regularizer which is adversarial in nature. We feed an input text sequence $x$ to the language model and extract the last hidden state representation of the model for $x$; denoted by $h_x$. $h_x$ is then fed to a  discriminator parameterized by $\theta_d$, which plays the role of an attacker who attempts to predict what the sensitive label (in our case, the author, $y$) for $x$ is. 
The output probability distribution of the discriminator for the input $h_x$, $p_d=\Pr(\cdot | h_x ; \theta_d)$ 
is then used to compute both the privacy loss $\mathcal{L}_{\textsc{LM-P}}$ of the language model and the discriminator loss $\mathcal{L}_{\textsc{D-CE}}$.
%
During training, the discriminator optimizes for better linking of the last hidden state representations to the authors.
Thus, the discriminator loss is $\mathcal{L}_{\textsc{D-CE}}(h_x,y;\theta_d)\!=\!  - \log \Pr(y | h_x ; \theta_d)$. 
Conversely, the language model optimizes $\theta_{lm}$ such that it (1) improves the utility of the language model and (2) flattens the probability distribution over authors.
Thus, we devise the following loss function:
%
%
\vspace{-1ex}
\begin{equation} \label{eq:adv-loss}
\begin{split}
    \mathcal{L}_{\textsc{LM}}(x; \theta_d, \theta_{lm})\!=\! \mathcal{L}_{\textsc{LM-CE}} + \lambda \mathcal{L}_{\textsc{LM-P}}
\end{split}
\vspace{-2ex}
\end{equation}
%
%
%
$\mathcal{L}_{\textsc{LM-CE}}$ is the utility loss, for which we use conventional cross entropy loss over the next-word predictions. $\mathcal{L}_{\textsc{LM-P}}$ is the \textit{privacy loss}:
\vspace{-1ex}
\begin{equation} \label{eq:adv-loss-pv}
\begin{split} 
\mathcal{L}_{\textsc{LM-P}}(h_x;\theta_d)=-\frac{1}{M} \sum_{c=1}^{M}\log \Pr(c | h_x ; \theta_d)
\end{split}
\vspace{-2ex}
\end{equation}
i.e. the KL divergence between the distribution over authors and the uniform distribution where $M$ is the number of classes (authors). 
%
The goal of this term is to drive the discriminator to predict randomly uniform outputs ~\cite{olympus}.
%
The reason we devised this loss as opposed to using $-\mathcal{L}_{\textsc{D-CE}}$ is that we do not just  want the discriminator to assign zero probability to the correct author, we want $p_d$ to be uniform so that it has no information about the correct author.
%
Hyperparameter $\lambda$ allows for trading off privacy and utility. 
%
%

\subsection{Triplet-based Loss Function}
One potential downside of the proposed adversarial regularizer is that the capacity of the discriminator must scale with the number of authors, and thus the size of the training data.
%
To better accommodate the larger number of authors in large datasets, we investigate another regularizer that does not require a discriminator.
We build on the intuition that to obfuscate an attribute, we can increase the distance between representations of samples that have the same label for that attribute while decreasing the distance between samples with different labels. 
To this end, we use the language model loss ($\mathcal{L}_{\textsc{LM}}$) of the previous section (Eq~\ref{eq:adv-loss}), and we set the privacy loss to be the triplet loss:
%
\vspace{-1ex}
\begin{align} \label{eq:priv-triplet}
\begin{split}
      \mathcal{L}_{\textsc{LM-P}} =
      \lVert h_x-h_p \rVert^2 - \lVert h_x-h_n \rVert^2
\end{split}
\end{align}

The triplet loss is commonly used in vision tasks for training embeddings that map images from the same category to neighboring points in the embedding space~\cite{triplet}.
We, however, invert this loss and use it for an opposite purpose: privacy regularization.
During the training of the language model, we select a ``baseline sample'', $x$, a positive sample $p$ (with different sensitive label)  and a negative sample $n$ (with the same sensitive label) and feed them through the language model and extract the last hidden states $h_x$, $h_p$ and $h_n$, respectively. 
%
%
We find the $l_2$ distance between $h_x$, $h_p$, and $h_n$ and based on their labels, add them to or subtract them from the loss.
%
To implement this, in practice, we sample a baseline batch and a second ``auxiliary'' batch during training. 
%
We feed both the baseline batch ($x$) and the auxiliary batch ($a$) through the language model and extract the last hidden states. 
We then calculate the distance between the last hidden states of the corresponding samples in the two batches.   %
If the samples have different labels for the sensitive attribute (author), we add their distance to the loss, otherwise, we subtract it. The privacy loss becomes: 
\begin{equation}
	\fontsize{10}{10}
\begin{split}
   \mathcal{L}_{\textsc{LM-P}} =  \sum_{\mathclap{i: y_{x_{i}} =    y_{a_i}}}^{}\lVert h_{x_i} - h_{a_i} \rVert ^2 
    -\sum_{\mathclap{j: y_{x_j} \neq y_{a_j}}}^{}\lVert h_{x_j} - h_{a_j} \rVert^2
\end{split}
\label{eq:loss}
\end{equation}

\vspace{-1ex}
\section{Evaluation}
\vspace{-1ex}

\begin{figure*}
    \centering
    \begin{subfigure}{0.3\textwidth}
     \centering
     \includegraphics[width=\linewidth]{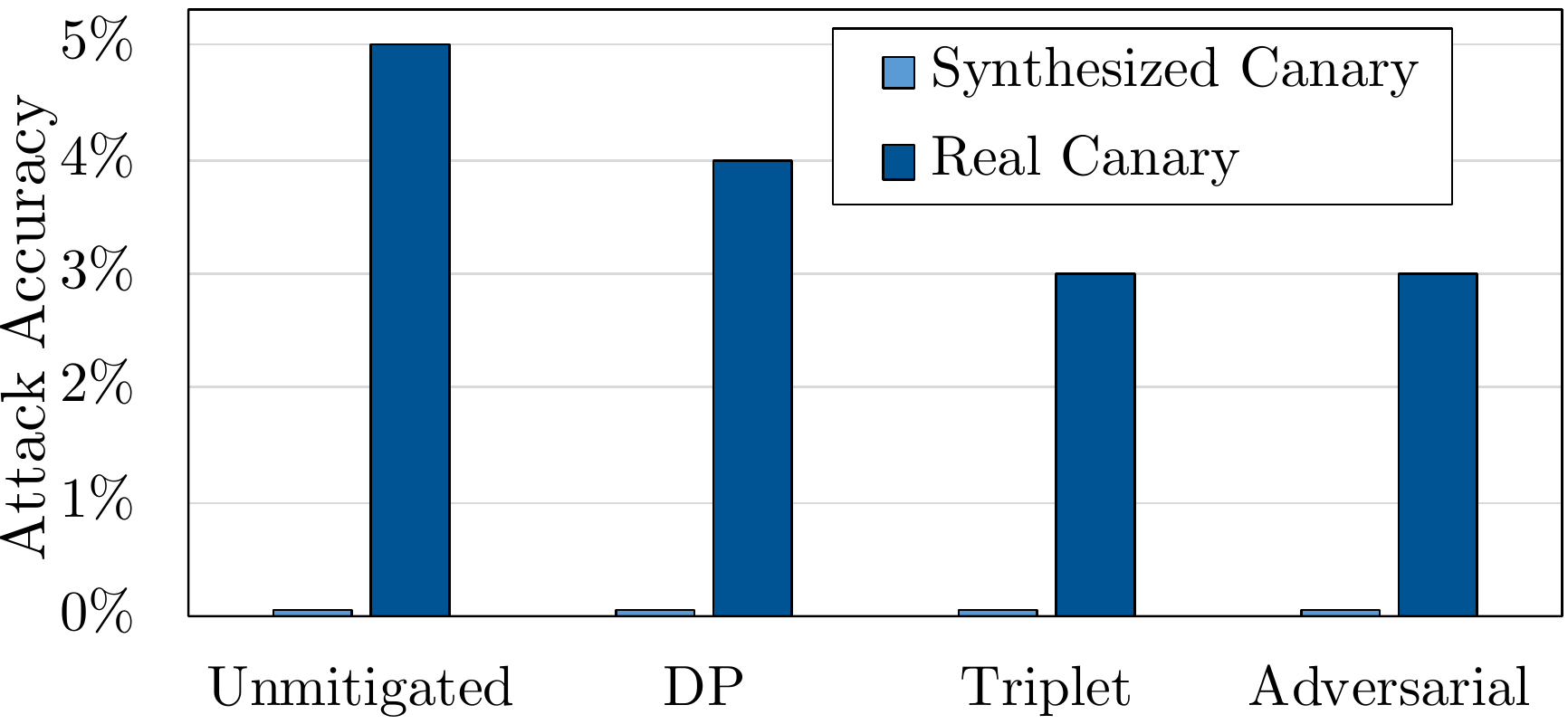}
     \caption{Avocado - High PPL ($\sim100$)}
     \label{fig:tab:high}
    \end{subfigure}
    \begin{subfigure}{0.3\textwidth}
    \centering
     \includegraphics[width=\linewidth]{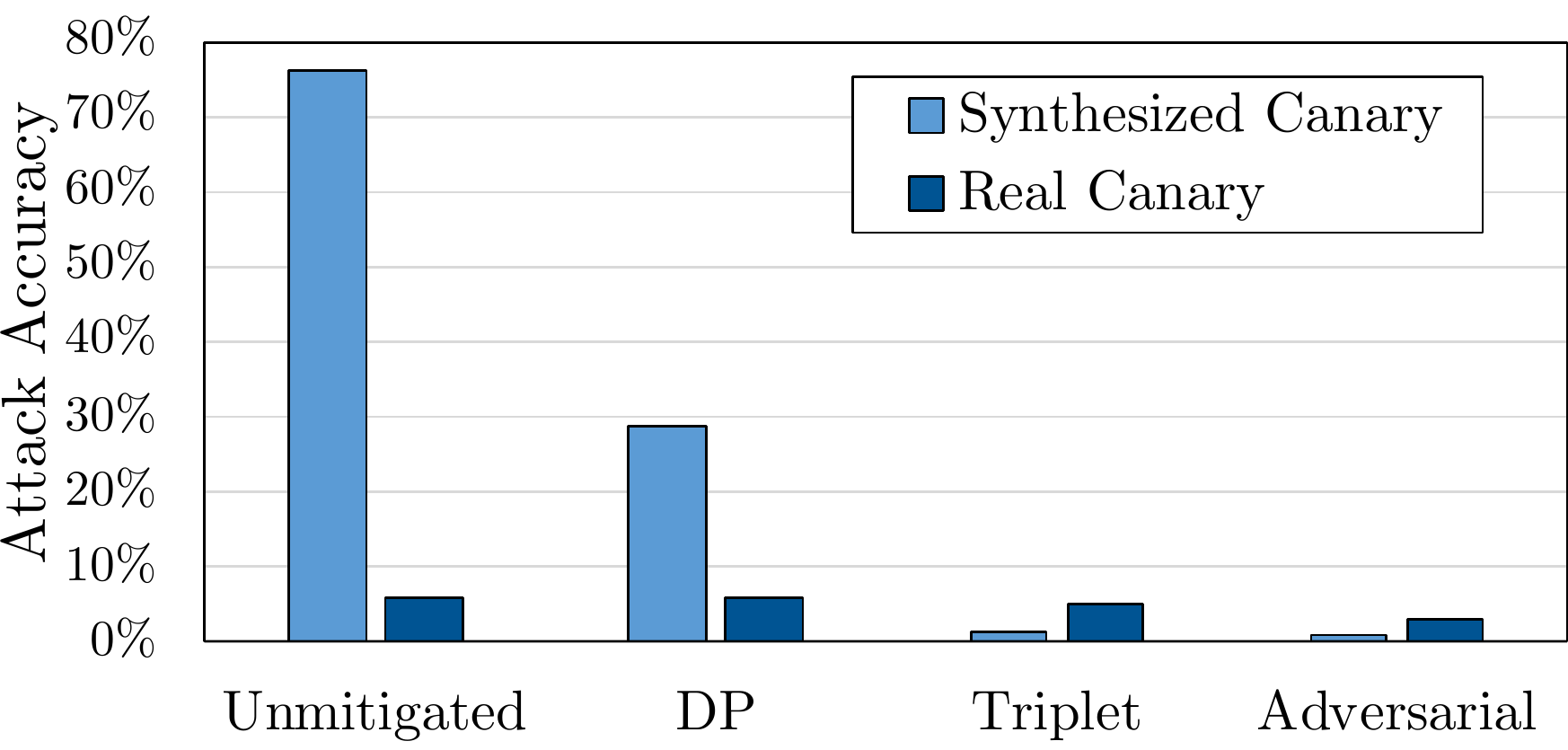}
     \caption{Avocado - Low PPL ($\sim60$)}
     \label{fig:tab:low}
    \end{subfigure}
        \begin{subfigure}{0.3\textwidth}
    \centering
      \includegraphics[width=0.95\linewidth]{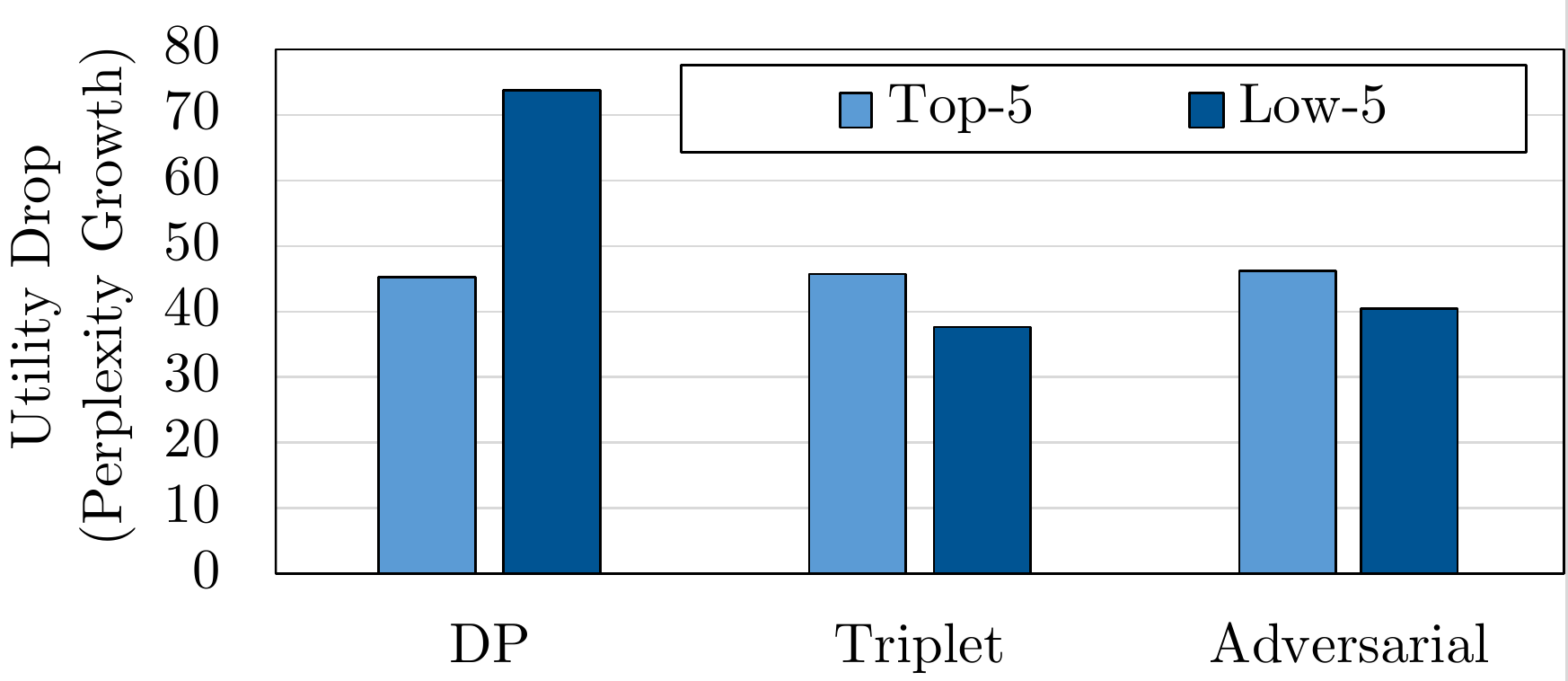}
     \caption{Impact on Different Subgroups}
     \label{fig:disparate}
    \end{subfigure}
        \caption{ (a, b) Tab attack results for reconstructing canary sequences for two utility levels. Higher attack accuracy indicates lower privacy. (c) Effect of different mitigations on utility of well represented (Top-5) and under-represented (Low-5) users for Avocado dataset. 
        }
        \vspace{-1ex}
    \label{fig:tab}
\end{figure*}



In our experiments, we use a subset of the Avocado corporate email dataset~\cite{avocado2015} with 100 users and 60,000 samples and a subset of Reddit dataset~\cite{volske-etal-2017-tl} with 10,000 users and 3 million samples. Both of these datasets are in English, covering formal and informal writing.
We create a $80/20\%$ training/test set split. 
We use a two-layer LSTM model as the language model for the next-word prediction task. 
%
We compare models trained with our proposed regularizer  to differentially private (DP) ones~\cite{abadi2016deep}. For the privacy accounting, we use Gaussian differential privacy~\cite{bu2019deep}.
We use language model perplexity as a measure of utility.
Due to space limitations, we focus evaluations on privacy metrics for several set levels of achieved test perplexity, listed in Table~\ref{tab:perps} in the appendix. 
%
See appendix~\ref{app:experiment} for a more detailed description of the experimental setup and extra analysis  of overheads and complexity of each regularizer.
%

\textbf{Privacy measurements w/ exposure metric.} 
To empirically compare the privacy of our methods to that of DP, we adopt the exposure metric introduced in ~\cite{carlini19}. 
The higher the exposure of a sequence, the more the model's memorization and the easier it is to extract the sequence from the language model. 
To measure exposure we insert sequences of five random words (canaries) to the training data (appendix~\ref{app:canaries}).
We insert unique canaries with different repetitions for each user, and measure the exposure of these canaries. 
%

Figure~\ref{fig:exposure} shows the exposure results per canary repetition. These results are averaged over all the users.
%
In each sub-figure, the perplexities of the models are similar, hence we can compare the privacy levels at similar utilities.
Fig.~\ref{fig:exposure:av-hi} compares trained models using different techniques on the Avocado dataset, where they all have relatively high perplexities compared to a fully trained conventional model (Table~\ref{tab:perps}).
Fig.~\ref{fig:exposure:av-lo} has the same setup, however, the models have lower perplexities. 
Naturally, for having better utility we are trading off privacy, which can be seen by comparing the exposure values in these two figures and observing that the second one has higher exposure values (lower privacy).
Finally, Fig.~\ref{fig:exposure:reddit} shows the exposure results for Reddit.
%

In all cases, we see that the unmitigated model has the highest exposure, as expected. We also observe that for canaries (patterns) that are repeated more than 9 times (for each user), our mitigation offers lower exposure compared to DP, especially in the high perplexity case. 
This is because clipping and noise addition in DP is attribute and data-agnostic, meaning that noise is added to all samples regardless of whether or not they contain sensitive information. Therefore, repeated patterns are less protected.
If we want to protect a pattern with $n$ repetitions, we would need to apply noise that is $n\times$ larger, which would degrade the utility gravely and would not yield the same perplexity.
For lower repetition canaries, our mitigations have comparable performance to DP.
For all these experiments the Gaussian differential privacy criterion $\mu$ is extremely large ($10^{20}$), which practically yields $\epsilon \sim \infty$. 
We also experimented with lower $\epsilon$ values (e.g. $\epsilon \sim 7$), however, it yields a model with perplexity of 650, having an extremely low utility. 

%
%
%
\textbf{Privacy measurements w/ tab attack accuracy.} 
In this experiment, we input the first token, and see if the entire sequence is reconstructed using the language model.
We report the rate of correct reconstruction of canaries as the accuracy of the attack. 
%
%
We 
use the synthetic canaries from the previous experiment, and also select ``real canaries'' from the training corpus to create a real-world scenario.
%
%
%
Fig.~\ref{fig:tab:high} shows that for a high perplexity model, the accuracy of the tab attack on synthesized canaries is very small, even for the unmitigated model. The unmitigated model reaches the designated perplexity in less than an epoch, and hence it does not memorize the canaries. 
For the real canaries, however, the memorization is higher, since they follow grammatical rules. 
%
In the lower perplexity case of Fig.~\ref{fig:tab:low}, we see that the synthesized canaries are mostly memorized by the unmitigated model. Our mitigations outperform DP, especially for the synthesized canaries. DP is not context-sensitive and applies the same amount of noise to all samples, thereby leaving correlated and higher repeated samples less-protected. 
Our mitigations, however, learn what sequences are link-able to their authors, and obfuscate them such that they no longer leak the ``identifying'' secret.

\textbf{Effect on under-represented users.}
Differential privacy has disparate impact on the accuracy of different subgroups of the dataset~\cite{disparate}. 
%
%
Here, we want to measure the effect of our mitigations on the utility of the model among users with various data samples.
For each user, we measure the average perplexity of the model for their samples in the test set, and then subtract this from the same value for an unmitigated model. This would yield the average drop in utility, per user. 
We compare the utility drop of well-represented users to under-represented ones by taking the top 5 users with the most samples and the bottom 5 users with the fewest samples from Avocado dataset.
We then measure the average utility drop over each group of 5 users on the test set.
Figure~\ref{fig:disparate} shows these results. 
We see that differential privacy has disparate impact, 29 points, on the two sub-groups of users (authors), %
whereas this gap is only 7 points for models trained with our mitigations.

It's important to remember that in general, distinguishing ``under-represented'' users from those whose data is similar to others but who have contributed fewer samples is a difficult task.
However, for Figure~\ref{fig:disparate}’s results, if these users’ data came from the same distribution as the ones with lots of samples (i.e. if these people were merely less-contributing), the utility loss would be similar for all groups when applying ``user-level'' DP (what we use).  DP’s disparate impact on the utility loss for these two groups suggests that, in our case, the less-contributing authors are probably also under-represented.
%
%
%


\vspace{-1ex}
\section{Conclusion}
\vspace{-1ex}
This work introduces two privacy mitigation methods to jointly optimize for privacy and utility. Extensive experiments show that our approach provides comparable and in certain cases a higher level of privacy compared to differentially private model training. 
We further empirically demonstrate, that our methods do not exhibit disparate impacts on under-represented groups and have significantly less overhead on training performance.

\section*{Ethical Considerations}
%
The Avocado corpus is licensed for research applications under strict terms intended to protect the privacy of the correspondents. 
While the end-user license agreement does not indicate what consent was granted by the participants, one term of the license is that “End user will obtain whatever training and approval is required by their organization for working with human subjects data”, which we have obtained (more details in~\cite{avocado2015}). 
While handling sensitive email data (Avocado) we made sure to abide by the terms of its end-user license agreement (EULA) which has provisions to protect the privacy of members of the corpus.
Furthermore, we took measures such as scrubbing named entities before using the data for model training.  The over-arching goal of our work is to contribute to language model development that protects the privacy rights of users who contribute their data.
%
While we rigorously evaluated our models by applying state-of-the-art attacks, deploying these models in real-world setups requires further verification that users’ privacy is preserved.
\section*{Acknowledgments}
The authors would like to thank the anonymous reviewers and meta-reviewers for their helpful feedback. We also thank Peter Kairouz and Mohammadkazem Taram for insightful discussions. Additionally, we thank our MSR colleagues and UCSD Berg Lab for their helpful comments and feedback. 

\bibliographystyle{acl_natbib}
\bibliography{anthology,custom}

\appendix
\captionsetup{belowskip=0pt}
\captionsetup{skip=0pt}
\setlength{\belowcaptionskip}{2pt}
\renewcommand{\baselinestretch}{1} 
\setlength{\textfloatsep}{3pt}%

\newcommand{\invis}[1]{}

\section{Appendix}
%
%
%
%
%

\subsection{Language Models} \label{app:lm}
Language models assign a probability distribution over a sequence of words. A statistical model for a sequence of words $x_1 \ldots x_n$ can be represented by the product of the conditional probability of the next word given all the previous words as $\Pr(x_1 \ldots x_n) = \prod_{i=1}^{n} \Pr(x_i | x_1 \ldots x_{i-1})$. Here $\Pr(x_i | x_1 \ldots x_{i-1})$ denotes the probability of the occurrence of word $x_i$ given the previous word sequence $x_1 \ldots x_{i-1}$. Recurrent Neural Networks (RNNs) are widely used for this task \cite{Bengio2003ANP, mikolov10} as RNNs can process variable-length input by processing words one at a time, updating its internal state and predicting the next word sequentially. Therefore, such variable-length conditional distributions can be effectively estimated with RNNs. In this work we use LSTMs \cite{lstm} in our language model.

\subsection{Experimental Setup}\label{app:experiment}
We use a subset of the Avocado dataset~\cite{avocado2015} with 100 users and 60,000 samples. We also use a subset of Reddit dataset~\cite{volske-etal-2017-tl} with 10,000 users and 3 million samples. For both datasets, we fix the vocabulary to the most frequent 40,000 tokens in the training corpus, and we create a $80/20\%$ training/test set split. 
We use a two-layer LSTM model as the language model for the next-word prediction task.
We set both the embedding dimension and LSTM hidden-representation size to 550 (with 55 million parameters).
We use a feed-forward two fully connected layer neural network with hidden dimension of 1000 as the author discriminator for the adversarial training regularization scheme.

For optimization, we use the Adam optimizer with the learning rate set to 1e-3 and batch size to 100. We trained our models on a single Titan Xp GPU accompanied by 12GBs of RAM, and two Intel Xeon E5-2620 CPUs with 256 GBs of RAM. Table~\ref{tab:perps}  shows the perplexity of the models used in the evaluations.


\subsubsection{Batching Strategy} \label{app:batching}
Our mitigations can be implemented using different batching strategies during training: uniform batches that contain training samples from different users, or per user batches that contain training samples from only a given user. 
Through experimentation, we observed that the second method performs better and we present our results under the second batching strategy.
The better performance is due to the fact that grouping the same users together and increasing the probability of samples from same people being placed in corresponding positions in a batch, helps the discriminator (in the adversarial training scheme) learn user (author) patterns faster. It also helps the triplet-based scheme to more efficiently distance samples from the same user.
This scheme we increases the probability that the auxiliary batch is selected from the same user's data, compared to randomly selecting two uniform batches.
This helps better distribute one user's data in the embedding space.
%

\begin{figure}     
    \centering
     \includegraphics[width=0.85\linewidth]{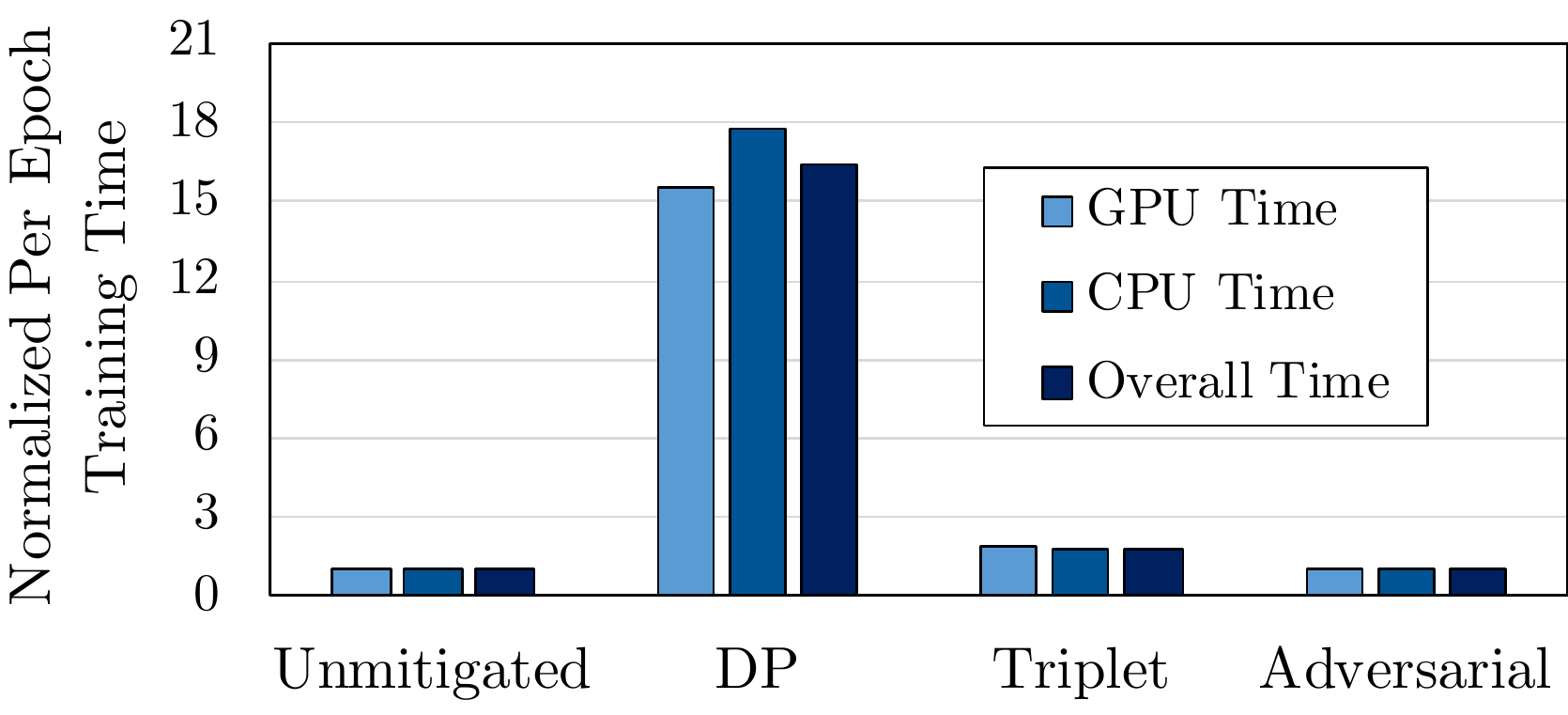}
        \caption{ Per epoch training time break down, normalized to conventional execution. Differential privacy is $16.44\times$ slower than conventional execution. Triplet and Adversarial are our proposed regularizations.} \label{fig:time}
\end{figure}
\begin{table}
\caption{Training and test perplexities of the models used in the evaluations.}
    \label{tab:perps}
    \resizebox{\linewidth}{!}{
  \centering
  \begin{tabular}{cl*{4}{c}}
    \toprule
    Dataset&& Unmitigated & DP & Adversarial (ours) & Triplet (ours) \\
    \midrule
    \multirow{1}{*}{{Avocado}} & Training & 93.5 & 94.2 & 104.7 & 101.4 \\ \cmidrule(lr){2-6}
    \multirow{1}{*}{{(High PPL)}}&Test & 103.5 & 93.5 & 96.6 & 96.5 \\ \midrule
    \multirow{1}{*}{{Avocado}} & Training & 36.8 & 63.8 & 56.7 & 54.8 \\ \cmidrule(lr){2-6}
    \multirow{1}{*}{{(Low PPL)}}&Test & 51.3 & 69.8 & 69.1 & 69.1 \\ \midrule
    \multirow{1}{*}{{Reddit}} & Training & 107.5 & 110.3 & 106.7 & 106.4 \\ \cmidrule(lr){2-6}
    \multirow{1}{*}{{}}&Test & 97.3 & 97.4 & 98.2 & 97.6 \\
    \bottomrule
  \end{tabular}
}

%

\end{table}

\subsection{Overhead and Complexity Analysis} \label{app:overhead}
\subsubsection{Training Time Measurements} \label{app:time}
\label{sec:profiler}
Here we compare the training time of our proposed regularizations to differential privacy. 
Figure~\ref{fig:time} shows the breakdown of the CPU and GPU (CUDA) time for our two proposed methods, and differential privacy, normalized to the conventional unmitigated execution time. 
The results show that differential privacy is extremely slower than our mitigations. Our adversarial training mitigation is overall only $1.06\times$ slower than conventional training, and our triplet-based mitigation is $1.80\times$ slower due to the need to process an auxiliary batch in each batch. 
The reason that our triplet-based mitigation is slower than the adversarial one is that the triplet based loss runs two batches through the language model during each iteration (the baseline batch and the auxiliary), whereas the adversarial training scheme runs only one batch of training data. 
Differential privacy, however, is $16.44\times$ slower than conventional execution, which is due to its per-sample gradient computation, which limits the possibility of parallelism.
It is noteworthy that in our experiments, we have applied the training time optimization suggested in~\cite{mcmahan2018general} for differential privacy, which increases parallelism through the use of ``micro-batches''. %
However, even with this optimization, we are still observing huge slow-downs.
%
%
%
Furthermore, differentialy private training of DNNs and RNNs takes exhaustive hyperparameter tuning to get the best privacy-utility trade-off, which is cumbersome~\cite{wu17, mirshghallah2020privacy}, and this slow training process makes the tuning of the parameters extremely harder.
%

\subsection{Added Parameters and Complexity}

Our \textbf{adversarial training} regularization includes an additional feed-forward discriminator DNN to predict the author of each sequence, as shown in Figure~\ref{fig:prop}. In our experiments, we use a feed-forward network with two fully connected layers and a hidden dimension of 1000. The input dimension is the same as the hidden state dimension of the language model (550) and the output dimension is the number of authors, 100 for Avocado, and 10,000 for Reddit dataset.
This means that the discriminator parameters scale-up with the number of authors. For the Avocado case, this would give $550\times1000 + 1000*100 = 650K$ extra parameters (compared to conventional training), and for Reddit it would be $550\times1000+1000\times10,000 = 10.55M$ parameters. This is comparatively reasonable considering the number of parameters in the language model (55M) as a tradeoff to improve the privacy of the model. It would be interesting to explore whether reducing the number of hidden dimensions of the discriminator as the number of users increase would be effective to balance the number of parameters our *adversarial method* adds to the parameters of the network. 

Our \textbf{triplet-based} regularization does not add any extra parameters to the model, which might be advantageous for settings with massive number of users. It does, however, need to feed an auxiliary batch to the language model, alongside the baseline batch which almost doubles the training time.

Figure~\ref{fig:time} and Section~\ref{app:time} show the break-down of the training time of DP training, our triplet-based method, and our adversarial method, compared to unmitigated training time.  Differential privacy does not add any additional complexity in terms of parameters. However, it adds computational complexity by having to compute the gradients of each sample separately, thereby not exploiting the batch processing parallelization offered by GPUs. This in turn slows down the training procedure extremely, making hyperparameter search infeasible for large networks.

\subsection{Exposure Metric}\label{app:canaries}
To empirically compare the level of privacy provided by our methods to that of a DP model, we adopt the exposure metric introduced in~\cite{carlini19}. 
This metric measures the extent to which a model memorizes samples in the training data. 
The higher the exposure metric for a sequence, the more the model's memorization and the easier it is to extract the sequence from the language model through text generation algorithms. 
To measure exposure we insert canaries to the training data. Our canaries are sequences of five random words from the vocabulary.
We insert canaries with different repetitions for each person. 
For Avocado dataset, each user (author) is assigned 14 unique canaries, each of them repeated  $[1, 2, 3, 4, 5, 6, 7, 8, 9, 10, 20, 30, 40, 50]$ times. 
We insert canaries with repition to mimic ``secrets'' that belong only to a certain user, but might be repeated by that user in multiple emails/texts. 
This means each user is assigned and overall of $1+2+3+...+50=195$ canaries. This yields and overall of $195\times100=19500$ canaries injected to the data.
This data is then used to train the model.
Once the model is trained, we can then measure the exposure of a given canary, by finding the probability assigned to it by the language model, and then finding the rank of that sequence by by sorting all probabilities assigned to all possible sequences of the same length. ~\citeauthor{carlini19} provide a method to estimate the rank for each sequence without having to actually enumerate and feed all possible sequences to the model and actually ranking all of them. 
Once we have measured the exposure metric for all canary sequences, we can use them to evaluate the privacy. 
For Reddit dataset, each user is assigned 5 unique canaries, with repetitions of $[1, 2, 5, 6, 20]$ times. We use less canaries in the latter so as not to contaminate the dataset as the Reddit dataset has less samples per user. 
The ``real canaries'' introduced in the experiments are chosen by feeding all the training data through 
an unmitigated model, and selecting the sample with highest perplexity for each user. We hypothesize that these sequences are more likely to contain 
sensitive information.

\invis{
\subsection{Related Work}
It has been long recognized that training machine learning models on datasets containing personal data comes with privacy issues. Many recent research efforts examine the vulnerability of such models or propose mitigations. 

Our work is closely related to \cite{coavoux18} and \cite{li2018}. In \cite{coavoux18}, the authors consider an attack model where an attacker eavesdrops on the hidden representations corresponding to an input and tries to recover information about the input text, such as gender or demographic information about the author of a text. They use adversarial training as mitigation to reduce the performance of an attacker and the setting is considered for the privacy of a text provided by a user at inference-time. In contrast, our goal is to prevent the model from learning sensitive information in the training set and we use adversarial training in this framework. On the other hand, \cite{li2018} use adversarial training to protect private author attributes such as age, gender, and location applied to part-of-speech tagging and sentiment analysis. Removal of such attributes can provide better generalisation, and can provide better privacy especially when latent representations of the data are shared for processing. We note that this is orthogonal to our work in terms of privacy as our goal is to protect sensitive information in the training set being encoded by the model. Our work is also related to \cite{olympus} where an obfuscation mechanism is applied to sensor data to limit the risk of disclosing private user information. The setting is considered where user data is fed to third party apps. We instead investigate the setting of a model training with privacy-utility consideration.

The {exposure} metric is introduced in \cite{carlini19} to quantitatively assess the unintentional memorization phenomenon occurring in generative sequence models. We use this metric to evaluate the mitigation effects of our approaches and extend it with an additional sequence extraction metric to evaluate the privacy risks in realistic settings. 

Prior work has studied the vulnerability of machine learning models to membership inference attack \cite{shokri17, yeom18, song19, long18, truex18}. The goal is to determine if a particular data record (or more generally data of a given user) belongs to the training set of the model. The key point making such attacks successful is that the models behave differently on the data records of their training set compared to the unseen data (i.e. performing better, predicting more confidently or providing better ranks etc.). This is also an orthogonal line of work to our paper as our main interest is not mitigating towards reducing this behavioral difference. 

The main framework with theoretical guarantees for model privacy is the application of differential privacy (DP)~\cite{dwork2011differential} to model training. DP makes provable guarantees about the privacy of a stochastic function of a given dataset. Differentially private stochastic gradient descent (DP-SGD) has been developed and applied to deep learning \cite{song13, abadi2016deep}. While DP-SGD is a promising avenue for private model training, further research is needed to determine that DP-SGD can yield acceptable privacy-utility trade-offs as current deep learning models require high privacy budgets to achieve reasonable utility losses and this can in turn lead to successful privacy attacks \cite{jayaraman19, wang18, disparate}, undermining the efficacy of the privacy mechanism.
}

\end{document}